\documentclass[conference]{IEEEtran}
\IEEEoverridecommandlockouts
\usepackage{amsmath,amssymb,amsfonts}
\usepackage{algorithmic}
\usepackage{graphicx}
\usepackage{textcomp}
\usepackage{xcolor}
\usepackage{listings}
\usepackage{booktabs}
\usepackage{hyperref}
\usepackage{algorithm}
\usepackage{algorithmic}

\hypersetup{
    colorlinks=true,
    linkcolor=blue,
    filecolor=magenta,
    urlcolor=cyan,
    citecolor=blue,
    pdftitle={A Federated Random Forest Solution for Secure Distributed Machine Learning},
    bookmarks=true,
    pdfpagemode=UseOutlines,
}

\definecolor{codegreen}{rgb}{0,0.6,0}
\definecolor{codegray}{rgb}{0.5,0.5,0.5}
\definecolor{codepurple}{rgb}{0.58,0,0.82}
\definecolor{backcolour}{rgb}{0.95,0.95,0.92}

\lstdefinestyle{mystyle}{
    backgroundcolor=\color{backcolour},   
    commentstyle=\color{codegreen},
    keywordstyle=\color{magenta},
    numberstyle=\tiny\color{codegray},
    stringstyle=\color{codepurple},
    basicstyle=\ttfamily\footnotesize,
    breakatwhitespace=false,         
    breaklines=true,                 
    captionpos=b,                    
    keepspaces=true,                 
    numbers=left,                    
    numbersep=5pt,                  
    showspaces=false,                
    showstringspaces=false,
    showtabs=false,                  
    tabsize=4,
    frame=single,
    framerule=0pt
}

\def\BibTeX{{\rm B\kern-.05em{\sc i\kern-.025em b}\kern-.08em
    T\kern-.1667em\lower.7ex\hbox{E}\kern-.125emX}}
\begin{document}

\title{A Federated Random Forest Solution for Secure Distributed Machine Learning

\thanks{This work has received funding from the FCT (Foundation for Science and Technology) under unit 00127-IEETA. J.M.S has received funding from the EC under grant agreement 101081813, Genomic Data Infrastructure.}
}

\author{
\IEEEauthorblockN{Alexandre~Cotorobai}
\IEEEauthorblockA{\textit{IEETA/DETI, LASI} \\
\textit{University of Aveiro}\\
Aveiro, Portugal \\
alexandrecotorobai@ua.pt}
\and
\IEEEauthorblockN{Jorge~Miguel~Silva}
\IEEEauthorblockA{\textit{IEETA/DETI, LASI} \\
\textit{University of Aveiro}\\
Aveiro, Portugal \\
jorge.miguel.ferreira.silva@ua.pt}
\and
\IEEEauthorblockN{José~Luis~Oliveira}
\IEEEauthorblockA{\textit{IEETA/DETI, LASI} \\
\textit{University of Aveiro}\\
Aveiro, Portugal \\
jlo@ua.pt}
}
\maketitle

\begin{abstract}
Privacy and regulatory barriers often hinder centralized machine learning solutions, particularly in sectors like healthcare where data cannot be freely shared. Federated learning has emerged as a powerful paradigm to address these concerns; however, existing frameworks primarily support gradient-based models, leaving a gap for more interpretable, tree-based approaches.
This paper introduces a federated learning framework for Random Forest classifiers that preserves data privacy and provides robust performance in distributed settings. By leveraging PySyft for secure, privacy-aware computation, our method enables multiple institutions to collaboratively train Random Forest models on locally stored data without exposing sensitive information. The framework supports weighted model averaging to account for varying data distributions, incremental learning to progressively refine models, and local evaluation to assess performance across heterogeneous datasets. Experiments on two real-world healthcare benchmarks demonstrate that the federated approach maintains competitive predictive accuracy—within a maximum 9\% margin of centralized methods—while satisfying stringent privacy requirements. These findings underscore the viability of tree-based federated learning for scenarios where data cannot be centralized due to regulatory, competitive, or technical constraints. The proposed solution addresses a notable gap in existing federated learning libraries, offering an adaptable tool for secure distributed machine learning tasks that demand both transparency and reliable performance.
The tool is available at \url{https://github.com/ieeta-pt/fed_rf}
\end{abstract}


\section{Introduction}

In recent years, advancements in machine learning have significantly transformed several fields, enabling intricate analytics and insights previously considered unattainable previously~\cite{yang2019federated}. In healthcare specifically, machine learning allows researchers and clinicians to leverage large-scale patient data to improve diagnostics, treatment outcomes, and operational efficiency. Nevertheless, despite these advances, the full potential of machine learning remains constrained by stringent data privacy regulations, such as the General Data Protection Regulation (GDPR)~\cite{chalamala2022federated}, institutional competition, and restrictive legal frameworks. Consequently, vast amounts of valuable data remain locked within dormant silos across institutions, severely limiting collaborative research and innovation.

Centralized secure analysis methods, such as Trusted Research or Execution Environments (TREs/TEEs), have been developed to address privacy concerns by providing strictly controlled settings for unified data analysis \cite{sabt2015trusted}. These environments allow researchers to perform comprehensive analyses on centralized datasets, ensuring high standards of reproducibility and data quality. However, the reliance on centralization itself often poses significant practical barriers. Regulatory limitations, competitive concerns among institutions, and technical constraints make it difficult or even impossible for many organizations to transfer their sensitive data into such controlled environments. Thus, although valuable, TREs are often unsuitable for scenarios where data must inherently remain distributed.

Federated Learning (FL) emerges as a compelling alternative to overcome these challenges by allowing multiple institutions to collaboratively train machine learning models without directly sharing their raw data \cite{pmlr-v54-mcmahan17a}. Under FL, each institution locally trains models using its private datasets, and only the model updates or parameters are exchanged and aggregated centrally. While federated learning frameworks predominantly employ gradient-based deep neural networks, these techniques have certain limitations, particularly regarding computational demands and interpretability \cite{li2020federated}. Deep neural networks can often function as black boxes, making them less suitable for contexts requiring transparent decision-making or for institutions with limited computational resources.

Despite significant developments in federated learning, existing frameworks have largely overlooked the integration of interpretable tree-based methods such as Random Forests (RF) \cite{breiman2001random}. Random Forests remain widely favored in many healthcare and enterprise applications due to their inherent transparency, robustness against overfitting, and minimal need for hyperparameter tuning. Nonetheless, adapting Random Forests to federated environments introduces distinct challenges, notably in effectively aggregating decision trees trained separately at each institution and managing heterogeneous data distributions across different sites. The current scarcity of specialized support for federated Random Forest approaches highlights a significant gap in the landscape of privacy-preserving machine learning.

In this paper, we present an novel PySyft-native implementation of Federated Random Forest with data-volume weighting and warm-start capabilities. While existing federated learning frameworks like FedTree~\cite{fedtree} and NVIDIA FLARE~\cite{roth2022nvidia} have made progress on tree-based models, our approach uniquely leverages PySyft's secure computation architecture to enable privacy-preserving tree ensemble training without direct data sharing, while supporting heterogeneous data distributions through weighted aggregation mechanisms.

\section{Background}
\subsection{Machine Learning}
Machine learning revolves around the idea of building models that discern underlying patterns from data rather than relying on hand-coded rules. These models learn from labeled examples, develop a statistical representation of the problem, and then apply their learned knowledge to previously unseen inputs. Various approaches address different data structures and task requirements. Neural networks excel in pattern recognition but are hard to interpret and require significant tuning \cite{oh2019towards}. XGBoost, described by Chen and Guestrin \cite{chen2016xgboost}, incrementally refines its model by adding new trees that correct the mistakes of the preceding ones, typically yielding strong performance but requiring careful parameter adjustments to avoid overfitting. Random Forests, introduced by Breiman \cite{breiman2001random}, take a different route by training multiple decision trees in parallel and combining their outputs through a majority vote. This ensemble strategy helps prevent overfitting, provides noise resilience, and can indicate the importance of each feature in the data, making the model more interpretable. Such interpretability can be crucial for regulated fields, where it is essential to demonstrate how each decision was reached \cite{ribeiro2016should}. Tree-based methods often outperform neural networks in tabular or structured data \cite{grinsztajn2022tree}, aligning with their prominence in explainable AI research \cite{lundberg2017unified}, yet neural networks excel in unstructured data such as text, speech and images \cite{li2022neural}. In federated learning scenarios, Random Forests remain particularly appealing because they allow each data silo to train its own trees without sharing raw data, then aggregate those trees or their predictions to form a robust global model. 

\subsection{Federated Learning}
Federated Learning, first proposed by McMahan et al. \cite{pmlr-v54-mcmahan17a}, is a machine learning approach where multiple entities (clients) collaborate to train a global model under the orchestration of a central server, without sharing their local data.

The typical federated learning process begins by initializing a global model on the central server. The server then distributes the global model to selected clients. Each client trains the model locally on their private data and sends the updated models back to the server. The server aggregates the local models to form an improved global model. This process repeats for multiple rounds until convergence or for a fixed number of iterations. 

FL addresses most of the privacy concerns by keeping raw data on the client and only sharing the model parameters or updates. This approach has seen significant adoption in mobile applications \cite{bonawitz2019towards}, healthcare \cite{rieke2020future}, and financial services \cite{mammen2021federated} \cite{byrd2020differentially}, where data privacy is mandatory.

Federated learning introduces several challenges not typically encountered in centralized machine learning, including communication efficiency \cite{konevcny2016federated}, where the multiple rounds of communication between the server and clients can be costly in terms of bandwidth and latency, statistical heterogeneity \cite{li2020federated}, since client data distributions may differ significantly, leading to increased difficulties in model aggregation and convergence, systems heterogeneity \cite{bonawitz2019towards}, because clients may have varying computational resources and connectivity, and lastly, even though raw data isn't shared, model updates may leak information about the underlying data, leading to privacy and security concerns \cite{mothukuri2021survey}.

Several frameworks have been developed to support federated learning implementations \cite{kholod2020open}. TensorFlow Federated (TFF), an open-source framework by Google, is designed for federated learning research and deployment, with a primary focus on neural network models. Flower is a versatile federated learning framework that supports heterogeneous clients and integrates with various machine learning frameworks \cite{beutel2020flower}, including PyTorch and TensorFlow. PySyft is a Python library for secure and private data analysis, compatible with multiple frameworks such as TensorFlow, Scikit-learn and PyTorch, offering the possibility of combining federated learning with additional privacy-preserving technologies.
OpenFL \cite{reina2021openfl}, developed by Intel, offers a framework for federated learning with a focus on healthcare and industrial applications, supporting both deep learning and traditional machine learning models. FATE (Federated AI Technology Enabler) is an industrial-grade federated learning framework that supports various machine learning algorithms and provides comprehensive security mechanisms including secure multi-party computation and homomorphic encryption \cite{liu2021fate}. 

Each of these frameworks offers distinct advantages, but they predominantly focus on gradient-based optimization methods, providing limited support for tree-based models like Random Forests.

\subsection{Federated Random Forests}
Adapting Random Forests to the federated setting is a challenging task. Unlike gradient-based models, tree-based models don't naturally support averaging or aggregation~\cite{hauschild2022federated}. Several approaches have been proposed in the literature, including model averaging \cite{liu2022federated}, where predictions from models trained on different data partitions are combined, and tree ensemble aggregation \cite{xiang2024transfer}, which involves collecting trees from different clients to create a larger forest. Feature-distributed training is also another method in which trees are built using features available at different locations \cite{gu2023commute}, and lastly, secure multi-party computation \cite{byrd2020differentially}, which uses cryptographic techniques to train trees across distributed data.

Our open-source implementation primarily follows the tree ensemble aggregation approach, where trees trained on client data are collected and selectively combined to form a global forest, taking into account weighted aggregation based on client data distributions.

\section{Methods}

\subsection{Framework Selection}

After assessing multiple federated learning frameworks, including TensorFlow Federated and Flower, we selected PySyft to develop this tool due to its distinctive theoretical and practical advantages. Its core innovation is an inverted approach to remote computation: rather than bringing data to the code, PySyft encapsulates all computational instructions and dispatches them securely to where the data reside. By keeping data on-site, this paradigm effectively addresses the privacy–utility paradox that often constrains collaborative data analysis. PySyft's governance structure further bolsters trust by enforcing explicit permission channels that allow data owners to inspect and approve any proposed operations before they occur on protected information, a safeguard vital for cross-institutional collaborations.

PySyft also delivers technical versatility by supporting TensorFlow, PyTorch, and Scikit-learn, which avoids the architectural compromises or dependency conflicts that can arise in tree-based algorithm development. Because computational instructions are wrapped and sent directly to each site, differential privacy safeguards like noise injection or privacy budgeting can be implemented exactly where data interactions take place, minimizing the need for additional privacy infrastructure and mitigating risks such as model inversion or membership inference attacks, this is a key feature that will enable our tool to provide privacy and confidentiality to data. In contrast to frameworks that assume uniform participation or enforce standardized models across nodes, PySyft's adaptable design accommodates the inevitable heterogeneity of real-world federations. This flexibility proved crucial in navigating the diverse institutional requirements of our deployment, where technical, regulatory, and organizational constraints demanded a more nuanced approach than many alternative solutions provide.

\subsection{Architecture}
Our Fed-RF implementation follows a client-server architecture as illustrated in Fig. 1, where a central coordinator (client) manages the federated learning process and multiple data servers (datasites) host local data and perform local training.

\begin{figure}[h]
    \centering
    \includegraphics[width=0.5\textwidth]{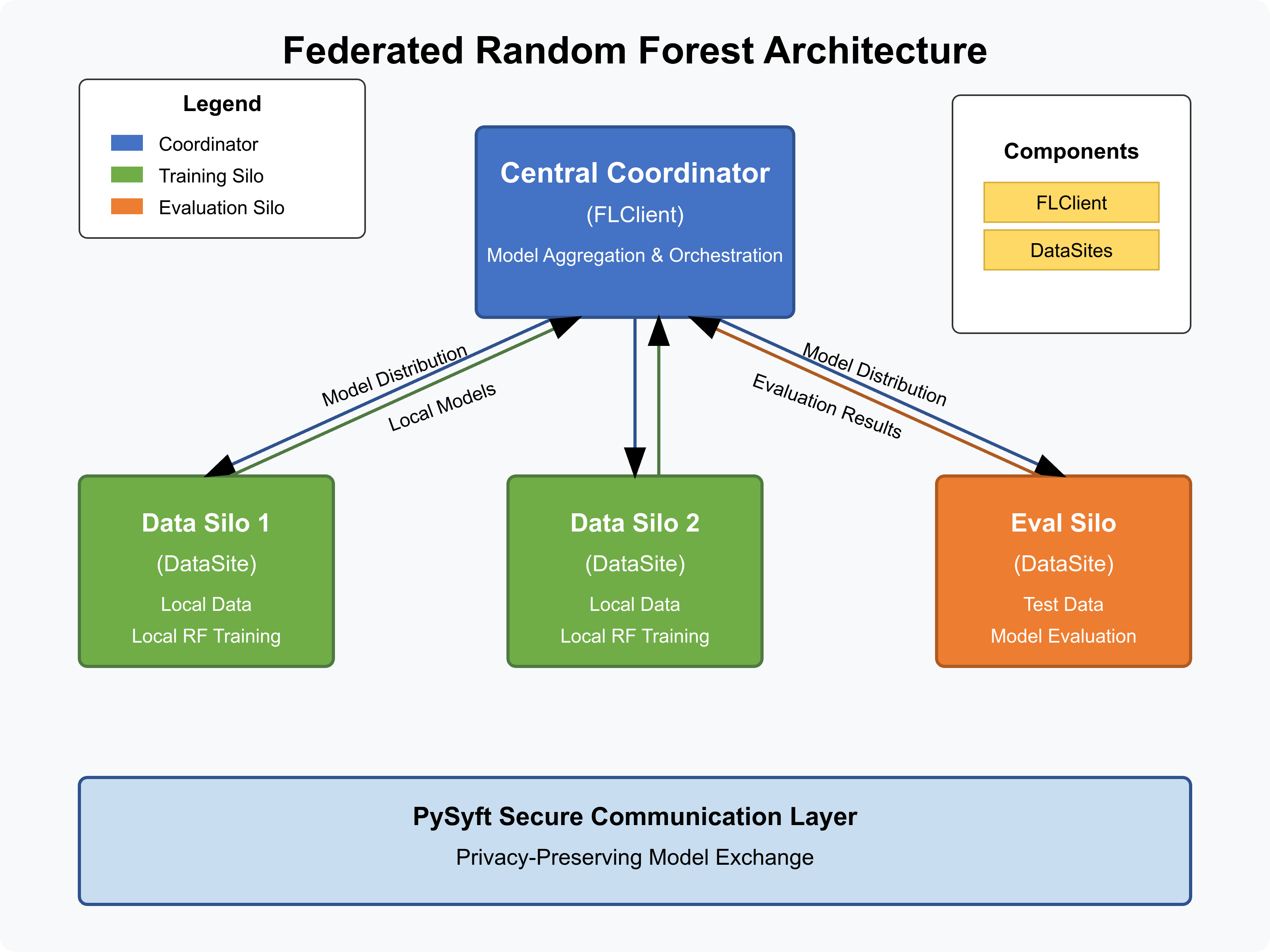}
    \caption{Fed-RF architecture}
    \label{fig:fed-rf-architecture}
\end{figure}

The implementation consists of two key components: the FLClient (central coordinator), which orchestrates the federated learning process, manages connections to data silos, and handles model aggregation and the DataSite, which represents a data silo with local data and compute capabilities

\subsection{Federated Learning Protocol}
Our federated learning protocol implements a multi-round process:
At initialization, the central coordinator connects to all data silos, sets data/model parameters and dispatches the training code to each silo.
Inside each silo a Random Forest model is trained on its local data and then models are serialized and sent back to the coordinator. After receiving all the locally trained models, it proceeds to model aggregation where trees from each model are aggregated using weighted sampling (ensuring the global forest has the same total number of trees as a single client model, composed of a mixture from all clients) based on silo importance and new global model is formed from the selected trees.

There is also the possibility for incremental learning where the global model is distributed back to silos for additional training. Each silo adds new trees to the existing forest using warm-start capabilities, and this process continues for multiple rounds, gradually improving the model. This approach is useful for handling scenarios where new data silos join mid-training.

\subsection{Weighted Model Aggregation}
A key innovation in our implementation is the weighted model aggregation strategy. Unlike neural networks, where weights can be directly averaged, Random Forests require a different approach. We have implemented two aggregation strategies: Uniform Sampling where trees are randomly selected from each client model with equal probability, and Weighted Sampling where trees are sampled with probability proportional to client-assigned weights \cite{luo2022tackling}. 
    
The weighting mechanism acknowledges that not all data silos contribute equally to the global model. Silos with larger or more representative datasets can be assigned higher weights, while those with smaller or niche datasets receive lower weights. This approach helps address the statistical heterogeneity challenge in federated learning. When client weights are not explicitly provided, our implementation can automatically distribute weights proportionally among clients.

\begin{algorithm}
\caption{Weighted Forest Aggregation}
\begin{algorithmic}[1]
\REQUIRE Parameters $modelP, dataP$, weights $w_1,...,w_n$
\ENSURE Global forest $F^*$
\STATE Dispatch training to each client in parallel
\STATE If any weights are missing, distribute equally or proportionally to successful clients
\FOR{each forest $F_i$ with weight $w'_i$}
    \STATE $k_i \leftarrow \lfloor w'_i \times N_i \rfloor$  \COMMENT{$N_i = |F_i|$ number of trees in $F_i$}
    \STATE Randomly select $k_i$ trees from $F_i$ and add to $T^*$
\ENDFOR
\STATE Create new global forest $F^*$ using combined trees in $T^*$
\RETURN $F^*$  \COMMENT{Time complexity: $O(N)$ where $N$ is total number of trees}
\end{algorithmic}
\end{algorithm}



\section{Results}
\subsection{Implementation and Package}
We have developed and released an open-source Python package called fed\_rf\_mk implementing our federated Random Forest approach. The package is available on GitHub and can be installed via pip:

\begin{lstlisting}[language=bash]
pip install fed_rf_mk
\end{lstlisting}

The implementation includes comprehensive documentation, example code, and utilities for setting up a federated learning environment.

\subsection{Experimental Setup}
To evaluate our federated Random Forest implementation, we conducted experiments with two real-world healthcare datasets: Dataset1, the AIDS Clinical Trials Group Study 175, and Dataset2, the Diabetic Retinopathy Debrecen dataset.

For each dataset, we established the following experimental scenarios. We used a centralized approach with a single Random Forest trained on the entire dataset as our baseline. We also tested a federated approach with data split across N silos with equal weight distribution. Additionally, we compared our results with current state-of-the-art solutions to provide context for the performance of our implementation. 

\subsection{Dataset 1: AIDS Clinical Trials}

The AIDS Clinical Trials Group Study 175 dataset (UCI ML Repository ID890) contains 2,139 records of HIV-infected adults from a randomized trial of antiretroviral therapies. Each entry includes 23 features—demographics, baseline clinical measures, treatment assignment, and medical history—with a binary outcome indicating death or censoring during the trial. The dataset supports classification tasks predicting mortality from baseline factors.

For context, we include the result from Malyala et al. \cite{malyala2024machine} on this dataset as the reference value, an accuracy of 0.8879 was achieved. The hyperparameter tuning conducted in that study identified the most fit parameters for this task, we will use them as the baseline for our experiment, being the number of estimators (4100) the most important parameter in this case study.
.

To evaluate federated learning on Dataset1, we partitioned the full dataset into \(N\) non‑overlapping silos (clients), where \(N\) varied from 1 to 10 to simulate different degrees of data fragmentation. Data was divided in a 20-80\% ratio, where 20\% of the data was assigned to the test silo, while the remaining 80\% was divided equally by the train silos. Each silo trained a Random Forest classifier locally using 4100 base estimators on its private data. After local training, the learned decision trees (model parameters) were securely transmitted to a central aggregator, which merged them into a single global model via simple parameter concatenation. We tested a second model with 2050 base estimators that was then incrementally updated with an additional 5 rounds of 410 estimators (totalizing 4100) to refine its predictive capacity before redistribution to all silos. Table \ref{tab:dataset1-table} presents the results of the experiments conducted with the AIDS Clinical Trials Group Study 175 dataset. The results are presented in terms of accuracy, precision, recall, F1 Score and Accuracy Deviation comparing with the centralized model.

\begin{table}[ht]
\centering
\caption{Results of the experiments conducted with the AIDS Clinical Trials Group Study 175 dataset}
\label{tab:dataset1-table}
\begin{tabular}{lccccc}
        \toprule
        \textbf{Mode} & \textbf{Accuracy} & \textbf{Precision} & \textbf{Recall} & \textbf{F1 Score} & \textbf{Acc Dev}\\
        \midrule
        Reference & 0.8879 & 0.9383 & 0.9157 & 0.9268 & - \\
        \midrule
        Centralized & 0.8808 & 0.8378 & 0.6138 & 0.7085 & 0\% \\
        3 Silos & 0.8785 & 0.8529 & 0.6105 & 0.7116 & 0.26\% \\
        5 Silos & 0.8552 & 0.8545 & 0.4947 & 0.6266 & 2.91\% \\
        10 Silos & 0.8088 & 0.9200 & 0.2421 & 0.3833 & 8.17\% \\
        \bottomrule
    \end{tabular}
\end{table}

\begin{figure}[h]
\centering
\includegraphics[width=0.5\textwidth]{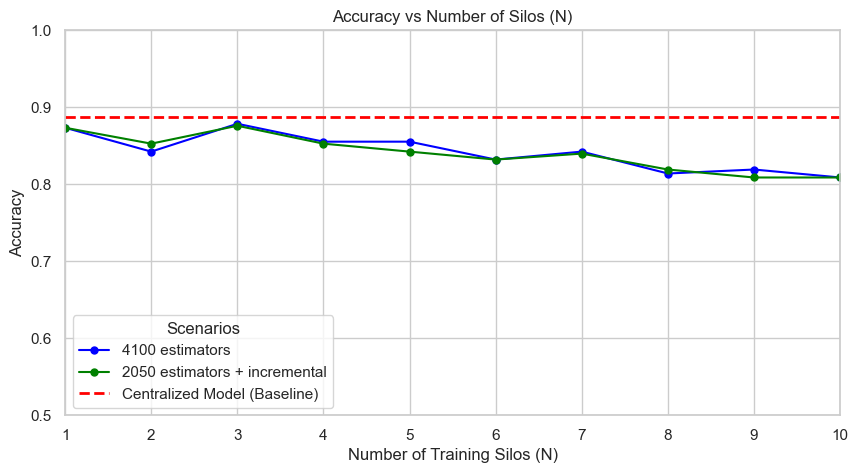}
\caption{Accuracy of different dataset distributions on dataset 1.}
\label{fig:dataset1-lineplot}
\end{figure}

The plot in Fig. \ref{fig:dataset1-lineplot} shows the accuracy of different dataset distributions.

\subsection{Dataset 2: Diabetic Retinopathy Debrecen}

The Diabetic Retinopathy Debrecen dataset (UCI ML Repository ID329) includes 1,151 patient records with 19 continuous features from retinal images and a binary label for diabetic retinopathy. It serves as a benchmark for automated diabetic eye disease screening.

We applied the same federated learning protocol described as in the previous Dataset. Each silo trains a Random Forest classifier initialized with 100 base estimators and a 50 base estimators model for incremental learning of 5 epochs of 10 additional estimators. We varied the number of silos \(N\) from 1 to 10 and evaluated performance using accuracy as the primary metric. Table \ref{tab:dataset2-table} presents the results of our experiments with the Diabetic Retinopathy Debrecen dataset.

\begin{table}[ht]
\centering
\caption{Results of the experiments conducted with the Diabetic Retinopathy Debrecen dataset}
\label{tab:dataset2-table}
\begin{tabular}{lccccc}
        \toprule
        \textbf{Mode} & \textbf{Accuracy} & \textbf{Precision} & \textbf{Recall} & \textbf{F1 Score} & \textbf{Acc Dev}  \\
        \midrule
        Reference & 0.688 & 0.700 & 0.690 & 0.690 & -\\
        \midrule
        Centralized & 0.718 & 0.773 & 0.667 & 0.716 & 0\% \\
        3 Silos & 0.702 & 0.758 & 0.621 & 0.683 & 2.23\% \\
        5 Silos & 0.701 & 0.761 & 0.603 & 0.673 & 2.37\% \\
        10 Silos & 0.653 & 0.729 & 0.558 & 0.632 & 9.05\% \\
        \bottomrule
    \end{tabular}
\end{table}

\begin{figure}[h]
\centering
\includegraphics[width=0.5\textwidth]{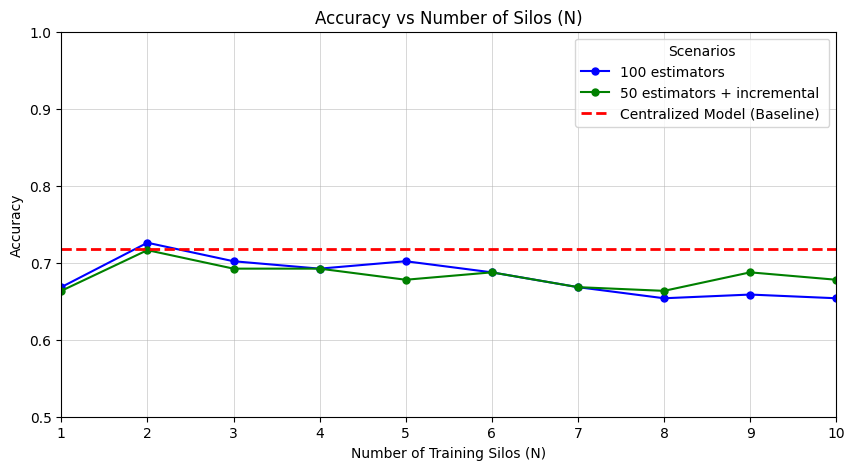}
\caption{Accuracy of different dataset distributions on dataset 2.}
\label{fig:dataset2-lineplot}
\end{figure}

The plot in Fig. \ref{fig:dataset2-lineplot} shows the accuracy of different dataset distributions.

\section{Discussion}

The experimental outcomes from our federated Random Forest implementation highlight key privacy–utility considerations when employing distributed machine learning. Tests across two distinct healthcare datasets show both promise and the inherent trade-offs of federated training.

In the AIDS Clinical Trials Group Study 175 dataset, our federated approach nearly matched the centralized baseline's accuracy of 0.8808, which is comparable to a state-of-the-art result of 0.8879 \cite{malyala2024machine}. As shown in Figure~\ref{fig:dataset1-lineplot}, performance decreased gradually with more extensive data fragmentation, reflecting the tension between preserving patient privacy and contending with statistically lean local datasets. Nonetheless, these results confirm that, under moderate distribution, federated methods can preserve much of the predictive capability observed in centralized models.

Turning to the Diabetic Retinopathy Debrecen dataset, our reference values were drawn from \cite{pragathi2022effective}, and the centralized model recorded 0.718 accuracy. An unexpected finding emerged when the (N=2) silos configuration surpassed that baseline with an accuracy of 0.7260. While likely driven by favorable data splits or initialization conditions, this observation hints that federated training can, under certain circumstances, yield performance that is on par with or even marginally exceeds centralized methods. A similar pattern was observed in the first dataset, where the (N=3) configuration also slightly peaked above the expected trend. However, as with the first dataset, results declined when the number of silos increased, dropping to 0.653 accuracy at 10 silos and highlighting that excessive fragmentation adversely affects model quality.

A comparison of the two datasets, one with a baseline near 0.88 and the other around 0.72, reveals that although the tasks differ in difficulty, the relative degradation pattern in federated settings remains consistent. This consistency demonstrates the generalizability of the federated Random Forest approach to different prediction problems, reinforcing its suitability for privacy-sensitive environments. The incremental learning capability, which allows for progressive model improvement through multiple federated rounds, represents a significant advantage for real-world deployments, allowing the global model to refine itself iteratively as new data or participating institutions joined.  Additionally, enabling each silo to locally evaluate the global model not only helps participants evaluate how well federated updates generalize to their unique data but also increases transparency, which is vital in clinical and regulatory contexts where trust in model decisions is fundamental. From a broader standpoint, our findings emphasize that federated Random Forest approaches can serve as a viable alternative to gradient-based frameworks, especially in domains requiring interpretability, modest computational resources, or straightforward model updating. Tree-based methods often align more naturally with clinical demands for transparent decision paths and explicit feature importances, and our experiments indicate that performance declines remain within a tolerable margin—typically below 9\%—relative to centralized baselines. This level of reduction is generally acceptable in sectors such as healthcare, where data security and regulatory compliance outweigh small gains in predictive performance. By retaining interpretability and safeguarding local data, our proposed system offers a sensible alternative to centralized methods in diverse, privacy-critical use cases, though organizations must still weigh whether slight drops in accuracy are justified by the corresponding advantages in privacy and transparency.

\section{Conclusion}

This work shows that Random Forest algorithms can be successfully adapted to federated learning with minimal accuracy trade-offs, particularly in privacy-sensitive domains where raw data cannot be centralized. By allowing each participant to retain complete ownership of their data and exchanging only model parameters, the proposed solution safeguards patient confidentiality while still enabling collaborative model development. The weighted sampling method for aggregating locally trained trees offers a robust way to account for heterogeneous datasets, ensuring that more representative silos exert appropriate influence on the final model. Additionally, support for incremental learning helps the system adapt over time, allowing new silos to join without disrupting ongoing training.
Evaluation on two real-world healthcare datasets indicates that although federated Random Forests experience some degradation relative to centralized training—ranging from 0.26–8.17\% in one dataset and 2.23–9.05\% in the other—these declines are typically acceptable in regulated environments where data security and explainability outweigh small gains in predictive performance. This PySyft-based open-source implementation provides an accessible option for organizations looking to deploy decentralized Random Forest models. Future research could refine aggregation strategies, incorporate more personalized learning, and explore differential privacy techniques to further narrow the accuracy gap and enhance protection against inference attacks. In doing so, tree-based federated learning can continue to mature as a viable alternative to gradient-based approaches when both interpretability and data privacy are fundamental.

\section*{Acknowledgment}
This work has received funding from the FCT (Foundation for Science and Technology) under unit 00127-IEETA. J.M.S has received funding from the EC under grant agreement 101081813, Genomic Data Infrastructure.

\end{document}